\documentclass[letterpaper, 10 pt, conference]{ieeeconf}  

\IEEEoverridecommandlockouts                              
\usepackage{times}
\usepackage{graphicx}
\usepackage{amsmath}
\usepackage{amssymb}
\usepackage{booktabs}
\usepackage{multirow}
\usepackage{xcolor}
\usepackage{url}
\usepackage{microtype}
\usepackage{hyperref}
\definecolor{darkred}{RGB}{150,0,0}
\definecolor{darkgreen}{RGB}{0,150,0}
\definecolor{darkblue}{RGB}{0,0,150}
\hypersetup{colorlinks=true, linkcolor=red, citecolor=blue, urlcolor=darkblue}

\newcommand{\pS}{p_S}
\newcommand{\pT}{p_T}
\newcommand{\DKL}{D_{\mathrm{KL}}}
\newcommand{\JSD}{D_{\mathrm{JSD}}}
\newcommand{\sg}{\mathrm{sg}}

\title{\LARGE \bf
On-Policy Distillation of Language Models \\for Autonomous Vehicle Motion Planning
}

\author{Amirhossein Afsharrad$^1$, Amirhesam Abedsoltan$^2$, Ahmadreza Moradipari$^3$, Sanjay Lall$^1$
\thanks{$^1$ Stanford University, $^2$University of California, San Diego $^3$University of California, Santa Barbara. {\tt \small afsharrad@stanford.edu } }
}

\begin{document}

\maketitle
\thispagestyle{empty}
\pagestyle{empty}

\begin{abstract}
Large language models (LLMs) have recently demonstrated strong potential for autonomous vehicle motion planning by reformulating trajectory prediction as a language generation problem. However, deploying capable LLMs in resource-constrained onboard systems remains a fundamental challenge. In this paper, we study how to effectively transfer motion planning knowledge from a large teacher LLM to a smaller, more deployable student model. We build on the GPT-Driver framework, which represents driving scenes as language prompts and generates waypoint trajectories with chain-of-thought reasoning, and investigate two student training paradigms: (i) on-policy generalized knowledge distillation (GKD), which trains the student on its own self-generated outputs using dense token-level feedback from the teacher, and (ii) a dense-feedback reinforcement learning (RL) baseline that uses the teacher's log-probabilities as per-token reward signals in a policy gradient framework. Experiments on the nuScenes benchmark show that GKD substantially outperforms the RL baseline and closely approaches teacher-level performance despite a 5$\times$ reduction in model size. These results highlight the practical value of on-policy distillation as a principled and effective approach to deploying LLM-based planners in autonomous driving systems.
\end{abstract}

\section{Introduction}
\label{sec:intro}

Motion planning is a cornerstone capability of autonomous driving systems, requiring a vehicle to generate safe, comfortable, and goal-directed trajectories in the presence of complex, dynamic environments. Classical approaches have relied on hand-crafted rules~\cite{treiber2000idm,urmson2008boss,moradipari2022predicting} or optimization-based methods, while modern learning-based planners leverage large-scale driving data to acquire planning behavior from human demonstrations~\cite{hu2022stp3,hu2023uniad}. Despite their empirical success, both paradigms suffer from limited generalization in long-tail scenarios and, in the case of neural planners, a lack of interpretability. A separate line of work focuses on ensuring safety in the generation of foundation models through rule-based methods~\cite{afsharrad2025multi,afsharrad2025cooperative,moradipari2020stage} and learning-based approaches~\cite{celestini2025generalizable,afsharrad2024convex}.

The emergence of large language models (LLMs) has opened a compelling new direction for autonomous driving. By virtue of pretraining on vast corpora, LLMs encode rich common-sense reasoning and world knowledge that is directly relevant to driving decisions. Researchers have begun applying LLMs across the autonomous driving stack, from scene understanding and question answering~\cite{cui2024survey,yang2023llm4drive} to high-level decision making~\cite{sha2023languagempc,fu2023drikeahuman} and end-to-end control~\cite{xu2024drivegpt4}. A particularly compelling direction is to reformulate motion planning itself as a language modeling problem: driving observations and ego states are converted to natural language prompts, and the LLM is trained to generate waypoint trajectories as structured text, optionally accompanied by a chain-of-thought reasoning trace~\cite{mao2023gptdriver}. This yields planners that are not only competitive with specialized neural methods~\cite{fazliani2025enhancing} but are interpretable by construction, providing human-readable explanations for every planned maneuver.

A central obstacle to deploying LLM-based planners in practice is their inference and computational cost~\cite{rathore2025turbocharging}. Modern capable language models are too large to run in real time on typical onboard automotive hardware. A natural solution is \textit{knowledge distillation}~\cite{hinton2015distilling}: train a smaller student model to replicate the behavior of a large teacher. However, standard supervised distillation suffers from a well-known \textit{train-inference distribution mismatch}. Because the student is trained on teacher-generated or ground-truth output sequences, the partial sequences it encounters at inference time (generated by its own imperfect policy) can differ substantially from those seen during training, causing compounding errors over the output sequence~\cite{ross2011dagger,agarwal2024gkd}. This problem is especially consequential for trajectory generation, where a coordinate error in an early waypoint could be propagated to all subsequent ones.

On-policy distillation addresses this problem by training the student on its \textit{own self-generated sequences}, using the teacher's token-level probability distributions as supervision on those on-policy samples~\cite{agarwal2024gkd}. This approach, formalized as Generalized Knowledge Distillation (GKD), has demonstrated strong results on language generation tasks including summarization, machine translation, and mathematical reasoning. Its deep connection to imitation learning~\cite{ross2011dagger} makes it particularly well-suited for sequential decision-making problems such as motion planning, where distribution mismatch has direct safety consequences.

In this work, we apply on-policy distillation to LLM-based autonomous driving. We train a Qwen3-8B teacher with supervised fine-tuning on the nuScenes-derived GPT-Driver dataset~\cite{mao2023gptdriver,caesar2020nuscenes}, and then distill its knowledge into a Qwen3-1.7B student using GKD. We compare this against a dense-feedback RL baseline that uses the teacher's log-probabilities as per-token reward signals within a policy gradient framework~\cite{zhao2026opsd}. Both methods train the student on its own on-policy rollouts, making their comparison a clean evaluation of full-distribution matching (GKD) versus sampled-token reward shaping (RL).

Our main contributions are as follows. First, we demonstrate that on-policy knowledge distillation can effectively compress an LLM-based autonomous driving planner, achieving near-teacher performance with a 5$\times$ smaller model. Second, we provide a principled and controlled comparison between GKD and a teacher-guided dense-feedback RL baseline, isolating the effect of the learning algorithm while keeping all other factors identical. Third, we show that the GKD student closely tracks the teacher on both trajectory accuracy and collision avoidance, and significantly outperforms the RL baseline on all metrics.

\section{Related Work}
\label{sec:related}

\subsection{LLM-Based Motion Planning}

GPT-Driver~\cite{mao2023gptdriver} is the closest precursor to this work. It reformulates motion planning as a language modeling problem, converts heterogeneous sensor observations and ego states into language prompts, and fine-tunes GPT-3.5 to generate waypoint trajectories alongside chain-of-thought reasoning. The model is evaluated on the nuScenes dataset and demonstrates strong imitation learning performance and few-shot generalization. A key insight of that work is that the GPT tokenizer naturally decomposes decimal coordinate values into integer and fractional parts, enabling hierarchical coarse-to-fine position estimation through the standard next-token prediction objective.

Our work inherits the task formulation and dataset from GPT-Driver but replaces the closed-source GPT backbone with the open-weight Qwen3 model family and focuses on teacher-student training rather than prompting or fine-tuning a single model. We note that since the original GPT-Driver work, language models have advanced substantially; the current frontier of proprietary models is represented by GPT-5 and similar systems. Our contribution is orthogonal to model capability: we study how to compress any capable teacher into a smaller student, regardless of the teacher's architecture.

Beyond GPT-Driver, a growing body of work applies LLMs to autonomous driving in various roles. DriveGPT4~\cite{xu2024drivegpt4} processes multi-frame video with a multimodal LLM for control prediction. LanguageMPC~\cite{sha2023languagempc} uses LLMs as high-level decision makers interfaced with a classical model predictive controller. Several works~\cite{fu2023drikeahuman,yang2023llm4drive,cui2024survey} explore how LLM reasoning and world knowledge can be grounded in driving affordances and scene representations. Our focus is distinct in that we study the \textit{compression} of an LLM planner rather than its application.

\subsection{Knowledge Distillation for Language Models}

Classical knowledge distillation~\cite{hinton2015distilling} trains a student to minimize the forward KL divergence between its token-level distributions and those of a teacher on a fixed dataset. Sequence-level KD (SeqKD)~\cite{kim2016seqkd} instead trains on teacher-generated output sequences, avoiding the need for the teacher to be available during training but sacrificing the rich soft-label signal. Both suffer from train-inference distribution mismatch for autoregressive models: the student learns on complete sequences but generates token-by-token at test time, causing errors to compound.

GKD~\cite{agarwal2024gkd} addresses this by training the student on its own generated sequences, using the teacher's token-level distributions as supervision on those on-policy samples. It allows flexible choice of divergence measure, including forward KL, reverse KL, and the generalized Jensen-Shannon divergence (JSD), each of which produces different quality-diversity tradeoffs. Recently, the OPSD work~\cite{zhao2026opsd} extends on-policy distillation to the self-distillation setting, where a single model acts as both teacher and student by conditioning on privileged ground-truth information. We adapt the standard GKD framework (with a separate teacher and student) to the structured output domain of autonomous driving planning.

\subsection{RL for Language Model Alignment and Distillation}

Reinforcement learning has been widely used to align language models with human preferences~\cite{ouyang2022instructgpt} and with verifiable reward signals~\cite{shao2024grpo}. In the context of distillation, teacher log-probabilities provide a natural dense reward signal: the advantage of a sampled token under the teacher relative to the student can be used directly in a policy gradient update. This approach, introduced in~\cite{zhao2026opsd} as an alternative to full-distribution GKD, provides on-policy supervision without requiring access to the full teacher vocabulary distribution at each step. It is closely related to the reverse-KL policy gradient used in MiniLLM~\cite{gu2024minillm}. We use this as our primary baseline, controlling for all experimental factors except the learning algorithm.

\section{Problem Formulation}
\label{sec:problem}

We follow the motion planning formulation of GPT-Driver~\cite{mao2023gptdriver}. At each planning step, the planner receives observations $\mathcal{O}$ and ego states $\mathcal{S}$ as inputs. The observations $\mathcal{O}$ encode the outputs of an upstream perception and prediction system: for each detected object in the scene, a natural language sentence describes its class, current position, and predicted future position. The ego states $\mathcal{S}$ encode the current velocity, acceleration, and heading angular velocity of the ego vehicle, as well as its historical trajectory over the past two seconds (four waypoints at 0.5-second intervals).

The goal is to produce a planned trajectory
\begin{equation}
    \mathcal{T} = \{(x_1, y_1), \ldots, (x_6, y_6)\},
\end{equation}
consisting of six waypoints at 0.5-second intervals over a 3-second horizon. The coordinate frame is ego-centric: the vehicle is located at the origin, the $y$-axis aligns with its current heading direction, and the $x$-axis is perpendicular. Coordinates are expressed in meters as decimal values.

Following GPT-Driver, the planner produces not only the trajectory but a full structured reasoning trace. The assistant output contains four components in order: (i) \textit{Notable Objects}, identifying the subset of perceived entities critical to the planned maneuver; (ii) \textit{Potential Effects}, describing when and how each critical object is predicted to influence the ego vehicle; (iii) \textit{Meta Action}, a high-level driving decision in natural language (e.g., ``TURN RIGHT WITH A DECELERATION''); and (iv) \textit{Trajectory}, the six numerical waypoints. This chain-of-thought structure, illustrated in Fig.~\ref{fig:prompt}, encourages the model to reason explicitly before committing to numerical coordinates, and provides interpretable explanations for every predicted maneuver.

Formally, let $x$ denote the language prompt encoding $(\mathcal{O}, \mathcal{S})$ and the high-level mission goal, and let $y^\star$ denote the ground-truth full assistant response. The planner is a conditional autoregressive language model $p_\theta(\cdot \mid x)$ trained to produce outputs $y$ that minimize displacement from $y^\star$. Importantly, the trajectory coordinates appear within the larger text output $y$, so the model must correctly generate the entire surrounding structure (section headers, reasoning text, and coordinate formatting) in addition to producing numerically accurate waypoints.

\begin{figure}[t]
    \centering
    \fbox{\parbox{0.95\columnwidth}{\small\vspace{0.3cm}
    \textbf{Prompt (excerpt):} \\
    Perception \& Prediction: car at (-8.67, 0.12), moving to (-8.50, -0.08). adult at (-1.21, 6.78), moving to (-1.29, 10.48). \\
    Ego-States: Velocity (vx,vy): (0.00, 1.46). Mission Goal: RIGHT \\[0.2cm]
    \textbf{Expected Output:} \\
    Thoughts: Notable Objects: adult at (-1.21, 6.78). Potential Effects: within safety zone at 1.0s. \\
    Meta Action: TURN RIGHT WITH A CONSTANT SPEED \\
    Trajectory: [(0.11,1.14), (0.45,2.28), (1.12,3.47), (2.18,4.54), (3.65,5.29), (5.49,5.58)]
    \vspace{0.3cm}}}
    \caption{Example input prompt and expected model output. The model generates a chain-of-thought reasoning trace before producing the final trajectory coordinates.}
    \label{fig:prompt}
\end{figure}

\section{Method}
\label{sec:method}

\subsection{Teacher Training}

We first train a strong teacher model using standard supervised fine-tuning (SFT). Concretely, we fine-tune Qwen3-8B~\cite{yang2025qwen3} on the GPT-Driver nuScenes training split~\cite{mao2023gptdriver,caesar2020nuscenes} using the \texttt{qwen3\_nothink} chat template, which disables the model's extended chain-of-thought thinking mode to produce deterministic, structured outputs. The teacher is trained to generate full planning responses including the reasoning trace, meta-action, and trajectory. All student experiments use a single fixed teacher checkpoint selected by validation performance.

The rationale for training the teacher with SFT rather than using an off-the-shelf pretrained model is that the nuScenes planning task requires a highly specific output format, domain-specific coordinate conventions, and the ability to reason about driving-specific entities. A general-purpose LLM would not reliably produce the structured output format required for trajectory parsing, making SFT on the task data a necessary prerequisite.

\subsection{On-Policy Generalized Knowledge Distillation}

\subsubsection{Motivation: Distribution Mismatch}

The core challenge in distilling an autoregressive planner is train-inference distribution mismatch~\cite{ross2011dagger}. In standard supervised training, the student is conditioned on ground-truth or teacher-generated prefix tokens $y_{<n}^\star$ when predicting token $y_n$. At inference time, however, the student must condition on its own previously generated tokens $\hat{y}_{<n}$, which may contain errors. Since autoregressive models predict each token conditioned on all previous ones, even small early errors can cascade: a slightly off first coordinate influences the distribution over subsequent coordinates, potentially causing the entire trajectory to drift.

This problem is particularly acute in motion planning. The six waypoints form a physically coherent trajectory, and the coordinate values span multiple orders of magnitude (centimeters to tens of meters). An error in the integer part of an early waypoint (e.g., predicting ``12'' instead of ``1'') corrupts the implicit representation of vehicle speed and direction that subsequent waypoints must be consistent with.

\subsubsection{GKD Objective}

On-policy GKD~\cite{agarwal2024gkd} resolves the mismatch by training the student on its own self-generated outputs. Given an input prompt $x$, the student samples a full response $\hat{y} \sim \pS^\theta(\cdot \mid x)$. The student is then trained to match the teacher's token-level distributions along this on-policy trajectory:
\begin{align}
\mathcal{L}_{\mathrm{GKD}}(\theta) &= \mathbb{E}_{x}\Bigl[\mathbb{E}_{\hat{y} \sim \pS^\theta(\cdot|x)}\bigl[D(\pT \| \pS^{\theta})(\hat{y} \mid x)\bigr]\Bigr], \label{eq:gkd}
\end{align}
where the token-averaged divergence is
\begin{align}
D(\pT \| \pS^{\theta})(\hat{y} \mid x) &= \frac{1}{|\hat{y}|}\sum_{n=1}^{|\hat{y}|} D\!\bigl(\pT(\cdot \mid \hat{y}_{<n}, x) \,\|\, \pS^{\theta}(\cdot \mid \hat{y}_{<n}, x)\bigr). \label{eq:tokendiv}
\end{align}
Crucially, gradients are not backpropagated through the student’s sampling process that generates the trajectory $\hat{y}$. The sampled prefixes $\hat{y}_{<n}$ are treated as constants, and only the token-level divergence in~(\ref{eq:tokendiv}) is differentiated with respect to $\theta$. This corresponds to ignoring the dependence of the trajectory distribution on the model parameters (i.e., dropping the score-function term), resulting in a biased but low-variance estimator that improves training stability and efficiency, similar to stop-gradient formulations in on-policy imitation learning~\cite{ross2011dagger}.


\subsubsection{Divergence Choice}

We use the generalized Jensen-Shannon divergence as $D$ in~(\ref{eq:gkd}):
\begin{equation}
\JSD^{(\beta)}(\pT \| \pS) = \beta \DKL(\pT \| m) + (1-\beta)\DKL(\pS \| m),
\label{eq:jsd}
\end{equation}
where $m = \beta \pT + (1-\beta)\pS$ is the mixture distribution and $\beta \in [0,1]$ interpolates between the forward KL ($\beta \to 0$) and the reverse KL ($\beta \to 1$). Forward KL is \textit{mode-covering}: it forces the student to assign probability mass wherever the teacher does, which can cause hallucination in low-capacity students. Reverse KL is \textit{mode-seeking}: it concentrates the student's mass on the teacher's highest-probability tokens, which can reduce diversity but improves output quality. JSD with $\beta = 0.5$ provides a balanced interpolation between these two behaviors. In our experiments we use the default TRL \texttt{GKDTrainer} parameters~\cite{trl}: $\beta = 0.5$ and a student data fraction $\lambda = 0.5$, meaning each training batch consists of 50\% on-policy student-generated sequences and 50\% ground-truth sequences.

\subsubsection{Why GKD Is Well-Suited for Planning}

The full-vocabulary supervision of GKD is especially valuable for coordinate generation. At each token position, the teacher provides a probability distribution over the entire vocabulary, effectively indicating which digit characters, decimal points, and delimiters are plausible continuations given the current trajectory prefix. This rich signal helps the student learn the implicit structure of coordinate sequences: that digits must form valid decimal numbers, that successive coordinates must encode physically realizable vehicle dynamics, and that the coordinate values must be consistent with the reasoning trace that preceded them. A scalar reward signal, by contrast, only tells the student whether the sampled token was relatively likely under the teacher, discarding all information about alternative continuations.

\subsection{Dense-Feedback RL Baseline}

As a baseline, we train a student using a teacher-guided policy gradient objective. This approach, introduced in~\cite{zhao2026opsd}, provides on-policy supervision using the teacher's log-probabilities as dense per-token rewards, without requiring access to the full teacher vocabulary distribution.

Given an input $x$ and a student rollout $\hat{y} \sim \pS^\theta(\cdot \mid x)$, the per-token advantage is defined as
\begin{equation}
A_n(x, \hat{y}) = \sg\bigl[\log \pT(\hat{y}_n \mid x, \hat{y}_{<n}) - \log \pS(\hat{y}_n \mid x, \hat{y}_{<n})\bigr],
\label{eq:advantage}
\end{equation}
where $\sg[\cdot]$ denotes the stop-gradient operation. The advantage $A_n$ is positive when the teacher assigns higher probability to the sampled token than the student does, and negative when the teacher assigns lower probability, providing a per-token signal about whether the student's choice was consistent with the teacher. The policy gradient objective is then
\begin{align}
\mathcal{L}_{\mathrm{RL}}(\theta) = -\mathbb{E}_{(x,y^\star)\sim\mathcal{S}}&
\left[\mathbb{E}_{\hat{y} \sim \pS(\cdot|x)}\left[\frac{1}{|\hat{y}|}
\sum_{n=1}^{|\hat{y}|} A_n(x,\hat{y})\right.\right. \notag \\
&\left.\left.\vphantom{\frac{1}{|\hat{y}|}\sum_{n=1}^{|\hat{y}|}}
\times \log \pS^\theta(\hat{y}_n \mid x, \hat{y}_{<n})
\right]\right].
\label{eq:rl}
\end{align}
The gradient of~(\ref{eq:rl}) takes the standard REINFORCE form: $A_n \nabla_\theta \log \pS^\theta(\hat{y}_n \mid x, \hat{y}_{<n})$, pushing the student's log-probabilities up on tokens the teacher preferred and down on tokens the teacher disfavored. No explicit KL penalty toward a reference policy is included, following~\cite{zhao2026opsd}.

\subsection{Comparison Between GKD and the RL Baseline}

Both methods generate on-policy student rollouts and use the teacher as the source of supervision. The critical distinction lies in the granularity of the learning signal at each token position $n$.

In GKD, the student receives feedback over the \textit{full vocabulary}: the JSD between the complete teacher and student distributions is minimized. This exposes the student to the teacher's probability mass over all plausible next tokens, including those not sampled in the current rollout.

In the RL baseline, the student receives feedback only at the \textit{sampled token} $\hat{y}_n$: the advantage $A_n$ provides a scalar signal about that one token, discarding all information about alternative continuations. This is analogous to the difference between a dense process reward and a sparse outcome reward in RL.

For coordinate generation in particular, the full-distribution signal of GKD can convey that, for example, the digit ``3'' and ``4'' are both plausible next tokens (corresponding to nearby valid coordinates), while ``9'' is implausible. The RL baseline, having sampled ``3'', only learns that ``3'' was slightly preferred by the teacher over the student's own estimate, with no information about ``4'' or ``9''.

\section{Experimental Setup}
\label{sec:experiments}

\subsection{Dataset and Evaluation}

We use the nuScenes autonomous driving dataset~\cite{caesar2020nuscenes} as processed by the GPT-Driver framework~\cite{mao2023gptdriver}. The dataset contains 1,000 driving scenarios covering diverse locations and weather conditions. We follow the official train/validation split, training all models on the training set and evaluating on the 5,119 validation frames of the official planner benchmark. Prompts are reconstructed from the raw nuScenes data using the original GPT-Driver preprocessing pipeline, ensuring exact comparability with prior work. All models generate outputs with greedy decoding and a maximum of 512 new tokens.

We report two families of evaluation metrics, both widely used in the autonomous driving planning literature:

\noindent\textbf{L2 displacement error (m).} The Euclidean distance between predicted and ground-truth waypoints, reported at 1, 2, and 3 second horizons and summarized under two averaging conventions. The STP-3 convention~\cite{hu2022stp3} computes a cumulative average: $\bar{L}_k = \frac{1}{k}\sum_{i=1}^{k} L_{0.5i}$ for horizon $k \in \{1,2,3\}$s, then averages across horizons. The UniAD convention~\cite{hu2023uniad} averages the exact-horizon L2 values at 1s, 2s, and 3s directly. Both are reported to provide a complete picture.

\noindent\textbf{Collision rate (\%).} The fraction of frames in which the ego-vehicle bounding box, placed at each predicted waypoint, overlaps with a ground-truth object bounding box. This measures trajectory safety independently of trajectory accuracy. Collision rates are reported at 1, 2, and 3 second horizons and averaged under the STP-3 convention.

We also report the \textbf{format error rate}: the fraction of examples for which the parser could not extract a valid 6-waypoint trajectory from the model output. Geometry metrics are computed only on successfully parsed examples, so format errors effectively count as missed predictions.

\subsection{Implementation Details}

All experiments are conducted on a single node of 8 NVIDIA H200 GPUs.

\noindent\textbf{Teacher.} Qwen3-8B~\cite{yang2025qwen3} fine-tuned using LLaMA-Factory~\cite{zheng2024llamafactory} with DeepSpeed ZeRO-3. Training uses learning rate $10^{-4}$, batch size 4 per device with 2 gradient accumulation steps (effective batch size 8), and the \texttt{qwen3\_nothink} chat template.

\noindent\textbf{GKD Student.} Qwen3-1.7B trained using TRL \texttt{GKDTrainer}~\cite{trl} with learning rate $5 \times 10^{-5}$, batch size 2 per device with 4 gradient accumulation steps (effective batch size 8), and maximum 512 new tokens per student rollout. Default TRL GKD parameters are used: $\beta = 0.5$ and $\lambda = 0.5$. The teacher's saved chat template is copied into the student training directory and used consistently across training, evaluation, and inference to ensure alignment between teacher and student tokenization.

\noindent\textbf{RL Baseline.} Qwen3-1.7B trained with the dense-feedback policy gradient objective in~(\ref{eq:rl}). Learning rate $5 \times 10^{-5}$, batch size 1 per device with 8 gradient accumulation steps (effective batch size 8). Student rollouts use temperature 0.7. All other settings match the GKD student.

\noindent\textbf{Checkpoint selection.} All three models are trained for 5 epochs with checkpoints saved after each epoch. We perform a sweep over all saved checkpoints on the validation set and report results from the best-performing checkpoint per model. This corresponds to epoch 3 for the teacher, epoch 3 for the GKD student, and epoch 1 for the RL student.

\section{Results}
\label{sec:results}

\subsection{Quantitative Comparison}

Table~\ref{tab:main} reports the main quantitative results on the nuScenes planning benchmark. The ordering Teacher $\geq$ GKD $\gg$ RL is consistent across all metrics.

\noindent\textbf{Trajectory accuracy.} The GKD student achieves an average L2 of 0.373\,m (STP-3) and 0.772\,m (UniAD), compared to the teacher's 0.355\,m and 0.730\,m. This represents a performance gap of only 5\% and 6\% respectively, despite the student having 5$\times$ fewer parameters (1.7B vs. 8B). The RL baseline, by contrast, achieves 0.579\,m (STP-3) and 1.092\,m (UniAD), which is 55\% and 41\% worse than the GKD student on the respective conventions. At the individual horizon level, the GKD student's advantage over RL grows with time horizon: at 1s the L2 ratio is approximately 1.9$\times$, while at 3s it is 1.4$\times$, suggesting that RL's errors compound more severely over longer output sequences.

\noindent\textbf{Collision rate.} The teacher achieves the best collision rates across all horizons. The GKD student follows closely, with an average STP-3 collision rate of 0.138\% versus the teacher's 0.101\%, while the RL student substantially lags behind at 0.363\%. The gap between GKD and RL is 2.6$\times$ on this safety metric, reinforcing that on-policy distribution matching produces trajectories that are both more accurate and safer than the sampled-token RL approach.

\noindent\textbf{Format reliability.} Both trained students produce zero format errors across all 5,119 validation examples, confirming that both learning algorithms reliably teach the structured output format. The teacher produces four format errors (rate ${\approx}0.08\%$).

\begin{table}[t]
\centering
\caption{Motion planning results on the nuScenes validation set (5,119 frames). L2 error (m, $\downarrow$) and collision rate (\%, $\downarrow$) reported at each horizon and as averages under STP-3~\cite{hu2022stp3} and UniAD~\cite{hu2023uniad} conventions. \textbf{Bold}: best per column. \underline{Underline}: second best. $^\dagger$Four of 5,119 examples failed to parse (${\approx}0.08\%$).}
\label{tab:main}
\setlength{\tabcolsep}{4pt}
\begin{tabular}{lccccc}
\toprule
\textbf{Method} & \textbf{L2 1s} & \textbf{L2 2s} & \textbf{L2 3s} & \textbf{STP-3} & \textbf{UniAD} \\
 & (m) & (m) & (m) & \textbf{Avg L2} & \textbf{Avg L2} \\
\midrule
Teacher       & \textbf{0.145} & \textbf{0.319} & \textbf{0.600} & \textbf{0.355} & \textbf{0.730} \\
GKD Student   & \underline{0.151} & \underline{0.334} & \underline{0.634} & \underline{0.373} & \underline{0.772} \\
RL Student    & 0.282 & 0.540 & 0.916 & 0.579 & 1.092 \\
\bottomrule
\end{tabular}

\vspace{0.4em}

\begin{tabular}{lcccc|c}
\toprule
\textbf{Method} & \textbf{Col. 1s} & \textbf{Col. 2s} & \textbf{Col. 3s} & \textbf{STP-3} & \textbf{Fmt.} \\
 & (\%) & (\%) & (\%) & \textbf{Avg Col.} & \textbf{Err.} \\
\midrule
Teacher      & \textbf{0.000} & \textbf{0.049} & \textbf{0.254} & \textbf{0.101} & ${\approx}0.08\%$$^\dagger$ \\
GKD Student  & \textbf{0.000} & \underline{0.068} & \underline{0.345} & \underline{0.138} & 0\% \\
RL Student   & \underline{0.049} & 0.274 & 0.765 & 0.363 & 0\% \\
\bottomrule
\end{tabular}
\end{table}

\subsection{Qualitative Comparison}

\begin{figure*}[t]
    \centering
    \includegraphics[width=1.8\columnwidth]{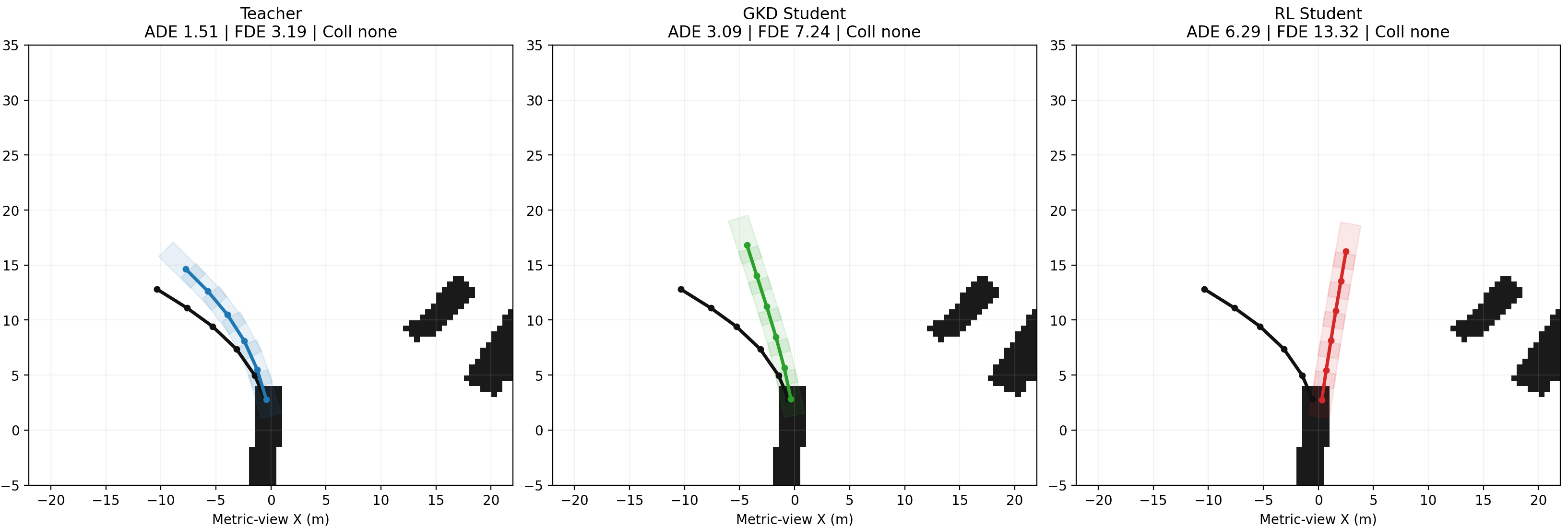}
    
    \caption{Qualitative trajectory comparison on a scenario where the ego 
vehicle executes a left turn. The ego vehicle is represented by the black 
box at the origin, and the other black boxes are the bounding boxes of 
surrounding agents detected by the perception system. The teacher (blue) 
correctly follows the ground-truth trajectory (black). The GKD student 
(green) misses the turn but stays closer to the ground truth 
(ADE~3.09\,m) than the RL student (red), which deviates most severely 
(ADE~6.29\,m vs.\ teacher ADE~1.51\,m). No collision occurs in any 
prediction.}
    \label{fig:qualitative}
\end{figure*}

Fig.~\ref{fig:qualitative} shows planned trajectories on a challenging right-turn scenario. The teacher correctly executes the turn, closely tracking the ground-truth trajectory. Both students miss the turn, predicting a largely straight trajectory instead; however, the GKD student stays substantially closer to the ground truth (ADE 3.09\,m) than the RL student (ADE 6.29\,m, a 2$\times$ larger error). This example illustrates how on-policy distribution matching helps the student better capture the teacher's turning behavior, even when it does not fully replicate it.

\subsection{Discussion}

\noindent\textbf{Full distribution vs. sampled token feedback.} The large performance gap between GKD and the RL baseline is consistent with findings in general language model distillation~\cite{agarwal2024gkd,zhao2026opsd}. Full-distribution matching at every token position provides a richer signal than per-token scalar reward shaping. In the motion planning context this difference is especially consequential: coordinate tokens form tightly constrained sequences where the teacher's full distribution encodes implicit knowledge about physically plausible vehicle dynamics, and the student benefits from seeing this complete distribution rather than a scalar advantage at one sampled value.

\noindent\textbf{Training stability and early stopping.} The RL baseline's best checkpoint occurs at epoch 1, with performance degrading in later epochs. This suggests overfitting or training instability characteristic of policy gradient methods when applied to structured sequence generation. The GKD student improves steadily through epoch 3, indicating more stable training dynamics. This is practically important: a method that is stable and predictable is easier to deploy in a safety-critical system.

\noindent\textbf{Parameter efficiency.} The GKD student achieves near-teacher performance with 1.7B parameters versus the teacher's 8B, a compression ratio of approximately 5$\times$. This level of compression, with only 5--6\% degradation in trajectory accuracy and competitive collision performance, suggests that on-policy distillation is a practical path to deploying LLM-based planners within the computational constraints of embedded automotive systems.

\section{Conclusion}
\label{sec:conclusion}

We have presented a study of knowledge distillation for LLM-based autonomous vehicle motion planning. Starting from a Qwen3-8B teacher trained on the nuScenes GPT-Driver benchmark, we distilled a 5$\times$ smaller Qwen3-1.7B student using on-policy generalized knowledge distillation, and compared it against a teacher-guided dense-feedback RL baseline under controlled conditions. The GKD student closely approaches teacher-level performance on trajectory accuracy and collision avoidance, while substantially outperforming the RL baseline on all metrics. These results demonstrate that on-policy distillation is a principled and practical approach to compressing LLM-based planners for deployment in resource-constrained autonomous systems.

Future work includes extending evaluation to closed-loop simulation, incorporating vectorized map and sensor inputs into the student prompt, studying the effect of the teacher-to-student capacity ratio on distillation quality, and exploring the integration of explicit safety objectives into the distillation training procedure.

\bibliographystyle{IEEEtran}
\bibliography{refs.bib}

\end{document}